%% file: main.tex
\documentclass{bmvc2k}

\usepackage{times}
\usepackage{epsfig}
\usepackage{graphicx}
\usepackage{amsmath}
\usepackage{amssymb}
\usepackage{booktabs}
\usepackage{multirow}
\usepackage{bm}
\usepackage{comment}
\usepackage{wrapfig}

\makeatletter
\newif\if@restonecol
\makeatother

\usepackage{algpseudocode}
\usepackage{amsmath}

\input{macrosjvg}

\begin{document}
\title{Push for Quantization: Deep Fisher Hashing }

\addauthor{Yunqiang Li$^\star$}{y.li-19@tudelft.nl}{1}
\addauthor{Wenjie Pei$^\star$}{wenjiecoder@outlook.com}{2}
\addauthor{Yufei zha$^\ast$}{zhayufei@126.com}{3}
\addauthor{Jan van Gemert}{j.c.vangemert@tudelft.nl
}{1}

\addinstitution{
Vision Lab, Delft University  of \\Technology, Netherlands
}
\addinstitution{
Tencent, China
}
\addinstitution{
School of Computer Science, \\ Northwestern Polytechnical \\University, Xi'an, China
}

\runninghead{Yunqiang Li et al.}{Push for Quantization: Deep Fisher Hashing}

\def\eg{\emph{e.g}\bmvaOneDot}
\def\Eg{\emph{E.g}\bmvaOneDot}
\def\etal{\emph{et al}\bmvaOneDot}

\maketitle

\begin{abstract}
Current massive datasets demand light-weight access for analysis. Discrete hashing methods are thus beneficial because they map high-dimensional data to compact binary codes that are efficient to store and process, while preserving semantic similarity. To optimize powerful deep learning methods for image hashing, gradient-based methods are required. Binary codes, however, are discrete and thus have no continuous derivatives. Relaxing the problem by solving it in a continuous space and then quantizing the solution is not guaranteed to yield separable binary codes. The quantization needs to be included in the optimization. In this paper we push for quantization: We optimize maximum class separability in the binary space. We introduce a margin on distances between dissimilar image pairs as measured in the binary space. In addition to pair-wise distances, we draw inspiration from Fisher's Linear Discriminant Analysis (Fisher LDA) to maximize the binary distances between classes and at the same time minimize the binary distance of images within the same class. Experiments on CIFAR-10, NUS-WIDE and ImageNet100 demonstrate compact codes comparing favorably to the current state of the art.
\end{abstract}

\section{Introduction}

Image hashing aims to map high-dimensional images onto compact binary codes where pair-wise distances between binary codes corresponds to semantic image distances,\emph{ i.e.,} Similar binary codes should have similar class labels. Binary codes are efficient to store and have low computational cost  which is particularly relevant in today's big data age where huge datasets demand fast processing.

\begin{wrapfigure}{r}{0.45\textwidth}
 \vspace{-20pt}
  \begin{center}
    \includegraphics[width=0.34\textwidth]{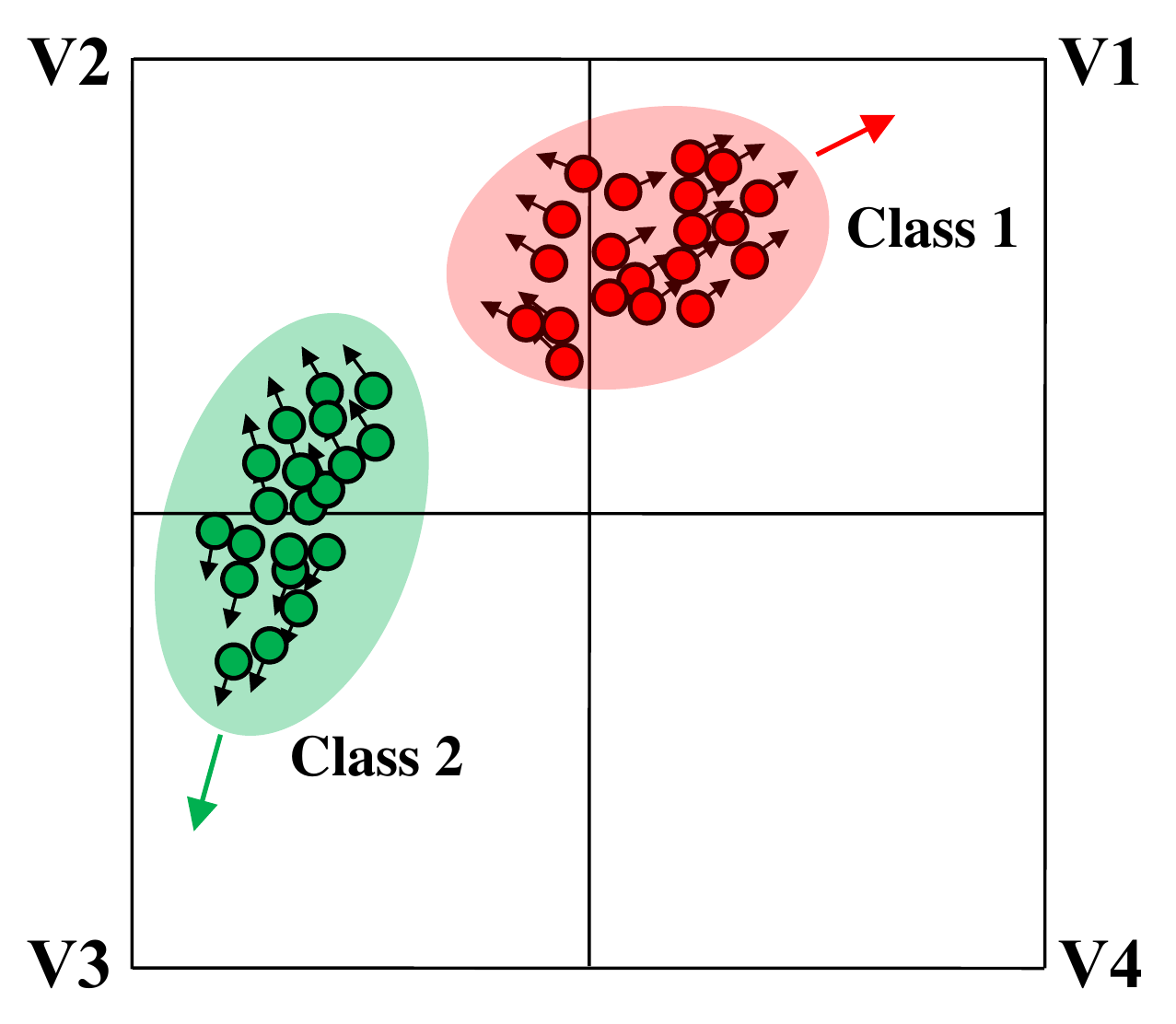}
  \end{center}
   \vspace{-20pt}
	\caption{ Example of two separable classes in a continuous space. After quantization (assign to grid cells)  the classes are no longer separable. In this paper we aim for separability  in the binary space. }
 \vspace{-10pt}	
\label{fig:vertexes}
\end{wrapfigure}

A problem in applying powerful deep learning methods for image hashing is that deep nets are optimized using gradient descent while binary codes are discrete and thus have no continuous derivatives and cannot be directly optimized by gradient descent. The current solution~\cite{ CaoCVPR18, Jiang_2017_CVPR, LiIJCAI2016, Liu2016DeepSH,  Zhang16CVPR,  Zhu:2016:AAAI} is to relax the discrete problem to a continuous one, and after optimization in the continuous space, quantize it to obtain discrete codes. This approach, however, disregards the importance of the quantization, which is problematic  because image class similarity in the continuous space is not necessarily preserved in the binary space, as illustrated in~\fig{vertexes}. The quantization needs to be included in the optimization.

In this paper we go beyond preserving semantic distances in the continuous space: We push for quantization by optimizing maximum class separability in the binary space. To do so, we introduce a margin on distances between dissimilar image pairs explicitly measured in the binary space. In addition to pair-wise distances, we draw inspiration from Fisher's Linear Discriminant Analysis (Fisher LDA) to maximize the binary distances between classes and at the same time minimize the binary distance of images within the same class

We have the following contributions. 1)  Adding a margin to pairwise labels pushes dissimilar samples apart in the binary space;  2) Fisher's criterion to maximize the between-class distance and to minimize the within-class distance leads to compact hash codes; 3) We show how to optimize this under discrete constraints and 4) We outperform state-of-the-art methods on two datasets, being particular advantageous for a small number of hashing bits.

\section{Related work}

\textbf{Amount of supervision.}
Existing hashing methods can be grouped on the amount of prior domain knowledge. Hashing methods without prior knowledge are applicable to any domain and include well-known methods such as Locality-Sensitive Hashing (LSH)~\cite{LSH1} and its extensions~\cite{Datar04localitysensitivehashing, Kulis09kernelizedlocalitysensitive, Kulis2009Fast,  Mu2010Non, Raginsky2009Locality}. If some knowledge about the data distribution is known in the form of an unlabeled training set, this knowledge can be advantageously exploited by unsupervised methods~\cite{Gong11iterativequantization, He2013K,Jiang2015Scalable, Kong2012Isotropic, Liu2014Discrete, Liu11hashingwith, Weiss2008Spectral} which learn hash functions by preserving the training set distance distribution. With the availability of additional prior knowledge about how samples should be grouped together, supervised methods~\cite{Chang2012Supervised, Gui2018Fast,  Lin2014Fast, Norouzi2011Minimal, Raziperchikolaei2015Optimizing, Shen2015Supervised, Zhang2014Supervised} can leverage such label information. Particularly successful supervised hashing methods use deep learning~\cite{LaiPLY15,Liu2016DeepSH, Liu17cvpr, Xia:2014:AAAI,  Yao:2016:IJCAI} to learn the feature representation. Supervision can be in the form of pairwise label information~\cite{CaoCVPR18, cao2017hashnet, Li2017NIPS,LiIJCAI2016,  Zhu:2016:AAAI} or in the form of class labels~\cite{Gui2018Fast, Li2017NIPS, Liu17cvpr, Shen2015Supervised}. In this paper we exploit both pairwise and class label knowledge, leading to highly compact and discriminative hash codes. \\

\noindent \textbf{Quantization in hashing.}
Several methods optimize the continue space and apply the \emph{sign} to obtain binary codes~\cite{ CaoCVPR18, Chang2012Supervised, Jiang_2017_CVPR, LiIJCAI2016, Liu2016DeepSH, Liu11hashingwith,  Zhang16CVPR,  Zhu:2016:AAAI, Zhao_2015_CVPR}.  A quantization loss is proposed in deep learning based hashing~\cite{CaoCVPR18, Jiang_2017_CVPR, Liu2016DeepSH, LiIJCAI2016, Zhu:2016:AAAI, Zhao_2015_CVPR} to force the learned continuous representations to approach the desired  binary codes. However, optimizing quantization alone may not preserve class separability in the binary space. An elegant solution is to employ \emph{sigmoid} or \emph{tanh} to approximate the non-smooth \emph{sign} function~\cite{cao2017hashnet, LaiCVPR2015}, but unfortunately comes with the drawback that such activation functions have difficulty to converge when using gradient descent methods. We circumvent these limitations by imposing the quantization loss in the discrete space, optimizing the  separability in the hashing space directly while guiding parameter optimization in the continuous space.

\noindent \textbf{Discrete optimization.}
Another branch of hashing methods to solve the discrete optimization is to utilize the class information to directly learn the hashing codes. For instance, SDH~\cite{Shen2015Supervised}, as well as its extensions such as FSDH~\cite{Gui2018Fast} and DSDH~\cite{Li2017NIPS}, propose to regress the same-class images to the same binary codes. While this kind of methods encourages a close binary distance between samples from the same class, they cannot guarantee the separability of samples from different classes.
In contrast, we propose to explicitly maximize the binary distances between classes and at the same time minimize the binary distances within the same class.

\section{Deep Fisher Hashing with Pairwise Margin}

\begin{figure}[t]
	\begin{center}
		\includegraphics[width=0.95\textwidth]{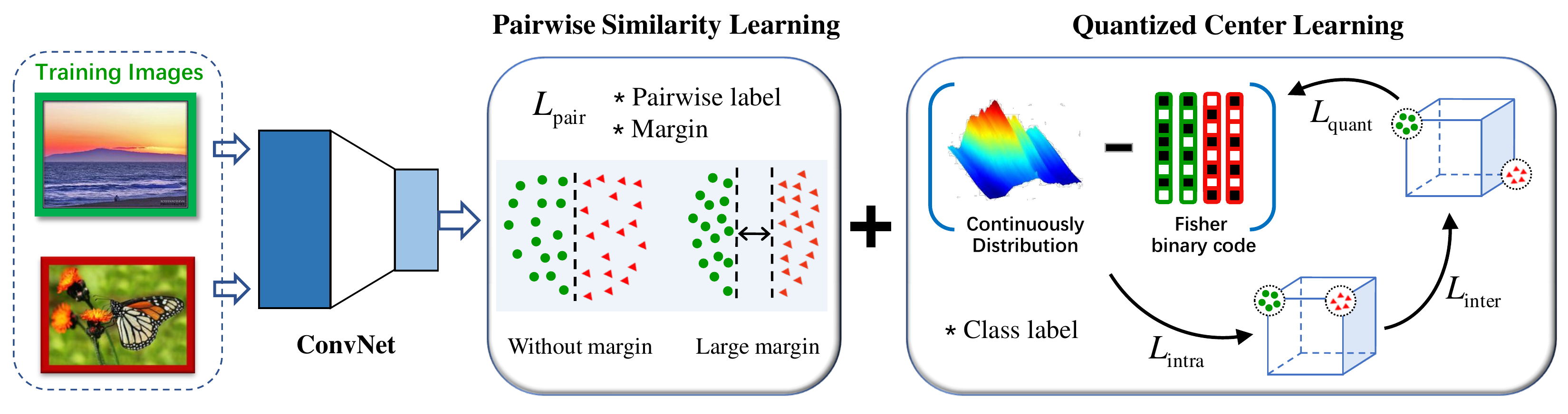}
	\end{center}
	\caption{  Images with class labels (red and green) are input to a CNN which outputs a $k$-dimensional continues representation $\mathbf{U}$. Module 1 maximizes a margin between dissimilar images in binary space ($L_{\text{Pair}}$). Module 2 minimizes binary distances within the same class ($L_{\text{Intra}}$)  and pushes different classes away ($L_{\text{Inter}}$) while quantizing $\mathbf{U}$ as binary codes ($L_{\text{Quant}}$).  }
	\label{fig:model} \vspace{-0.15in}
\end{figure}

In \fig{model} we illustrate our model.
Two components steer the discrete optimization: 1) A Pairwise Similarity Learning module  to preserve semantic similarity between image pairs while using a margin to push similar and non-similar images further apart ($L_\text{pair}$). 2) A Quantized Center Learning module inspired by Fisher's linear discriminant that maximizes the distance between different-class images ($L_\text{inter}$) whilst minimizing the distance between same-class images ($L_\text{intra}$) where the binarization requires minimizing quantization errors  $L_\text{quant}$.  These two modules are optimized jointly on top of a convolutional network (CNN).

For a train set of $N$ images 
$\mathbf{X} = \{ \mathbf{x_i}\}_{i = 1}^N$, with $M$ class labels $\mathbf{Y} = \{ \mathbf{y}_i \}_{i=n}^N \in \mathbb{R}^{M \times N}$, where $\mathbf{y}_i \in \mathbb{R}^M$  is a vector with all elements $\ge 0$ that sums to $1$, representing the class proportion of sample $\mathbf{x_i}$. For single-label (multi-class) $\mathbf{y}_i$ reverts to a one-hot encoding $\{0,1\}^M$. If $\mathbf{x}_i$ has $m$ multiple labels, each has a value of $1/m$ in $\mathbf{y}_i$. The last layer of the CNN $\mathbf{U} = \{\mathbf{u}_i\}_{i=1}^N \in \mathbb{R}^{K \times N}$ is the learned representations of $\mathbf{X}$. The output codes $\mathbf{B} = \{\mathbf{b}_i\}_{i=1}^N \in \{ -1, 1\}^{K \times N}$ are
the discretized binary values corresponding to $\mathbf{U}$ with each image encoded by $K$ binary bits.

\subsection{Pairwise Similarity Learning}

The main goal of hashing is to have small distances between similar image pairs and large distances between dissimilar image pairs in the binary representation. For binary vectors $\mathbf{b_i},  \mathbf{b_j} \in \{ -1, 1\}^K$,
the Hamming distance $D_H(\mathbf{b}_i, \mathbf{b}_j) =  \frac{1}{2}(K - \mathbf{b}_i^\intercal \cdot \mathbf{b}_j) = \frac{1}{4}D_E(\mathbf{b}_i, \mathbf{b}_j)$.
Since $K$ is a constant, it can be left out and we define the dissimilarity  $D(\mathbf{b}_i,\mathbf{b}_j)  = - \frac{1}{2} \left( \mathbf{b}_i^\intercal \cdot \mathbf{b}_j \right)$.
Note that larger dissimilarity $D$ indicates larger Hamming distance and less similarity.

\begin{wrapfigure}{r}{0.45\textwidth}
 \vspace{-4pt}
  \begin{center}
  \includegraphics[width=0.42\textwidth]{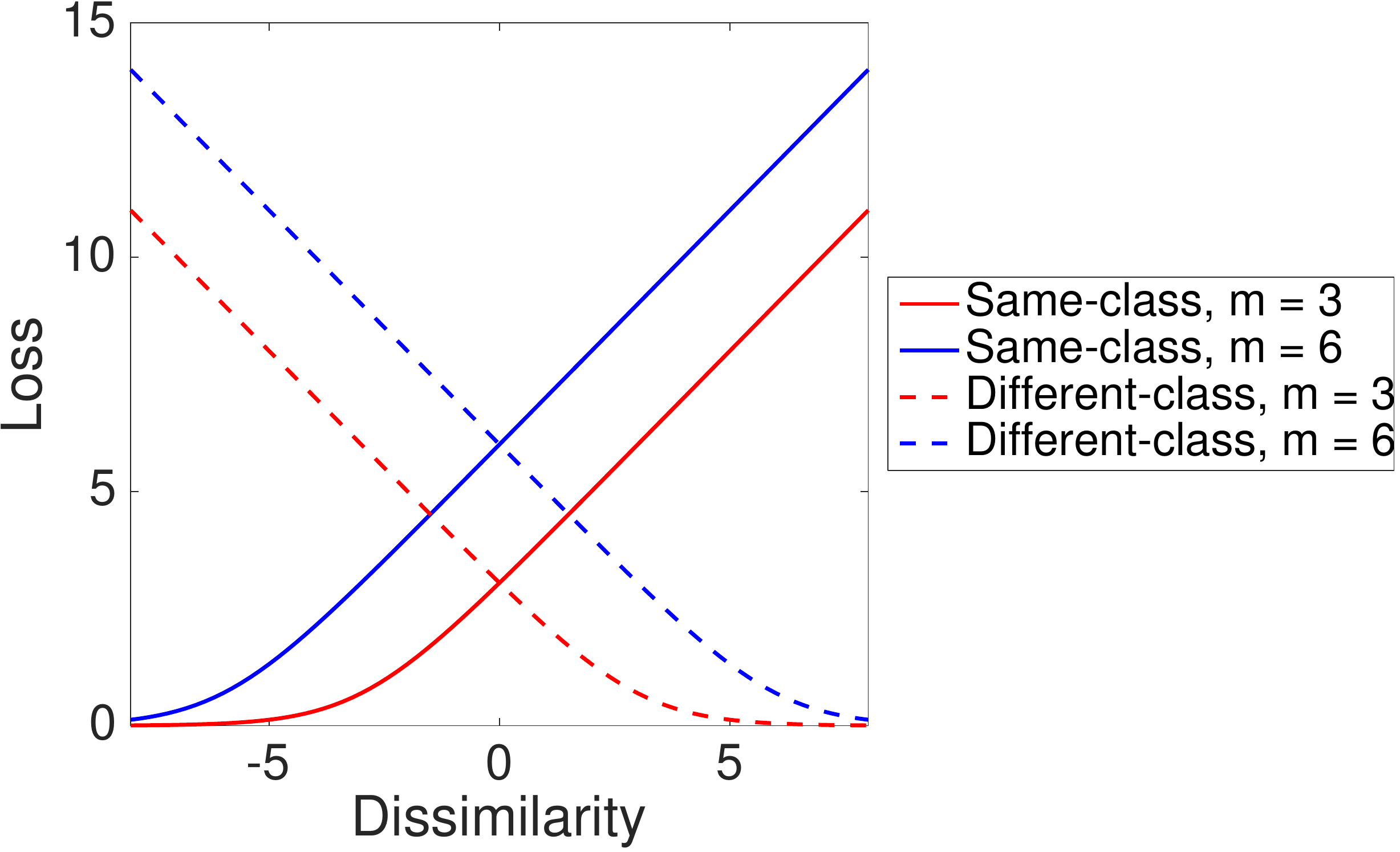} \\
  \end{center}
   \vspace{-15pt}
	\caption{ Our symmetric large margin logistic loss of both same-class and different-class cases as a function of the dissimilarity with different margin $m$. Larger $m$ encourages separation.}
 \vspace{-4pt}	
	\label{fig:loss}
\end{wrapfigure}
Similar images should share many binary values while dissimilar images should share few binary values.  Given the dissimilarity $D(\cdot,\cdot) \in (-\frac{1}{2}K, \frac{1}{2}K)$, a dissimilarity of 0 between binary vectors $\mathbf{b}_i$ and $\mathbf{b}_j$ means that half of their bits are different.  To encourage more overlapping bits for similar images and less overlapping bits for dissimilar images, we add a margin $m$ to a symmetric  logistic loss centered at 0:
\begin{equation}
L^S(D) = \log(1+e^{D+m});  L^D(D) = \log(1+e^{-D+m}).
\label{eqn:inner_positive_pro}
\end{equation}
The hyper-parameter $m \geqslant 0$ controls separation between similar pairs $S$ and dissimilar pairs $D$.  When $m = 0$, our model will turn into the classical way used in~\cite{Li2017NIPS, LiIJCAI2016}.
\fig{loss} illustrates the loss curves of same-class pairs and different-class pairs as a function of dissimilarity calculated by our dissimilarity measure with various values of $m$. Larger margin can help to pull same-class pairs together while push different-class pairs far away.

The Pairwise Similarity module minimizes the large margin logistic loss:
\begin{equation}
\begin{split}
 {L}_{\text{pair}}  & =   \sum_{(i, j) \in \mathcal{S}}  L^S ( D \left(\mathbf{b}_i, \mathbf{b}_j ) \right )
+ \sum_{(i, j) \in \mathcal{D}} L^D ( D  \left(\mathbf{b}_i, \mathbf{b}_j \right) ) \\
& {s.t.} \ \ \   \mathbf{b}_i, \mathbf{b}_j  \in \{-1, 1\}^{K}, \ \  i,j = 1,...,N.
\end{split}
\label{eqn:pair_Loss}
\end{equation}
Since $\mathbf{b}_i$ and $\mathbf{b}_j$ are discretized hashing codes from the continuous output of the CNN ($\mathbf{u}_i$ and $\mathbf{u}_j$), thus it is hard to back-propagate gradients from $L_{pair}$ to parameters of the CNN.
 To make the CNN trainable with $L_{pair}$,
 we introduce an auxiliary variable $\mathbf{u}_i = \mathbf{b}_i$.
Then we apply Lagrange multipliers to get the Lagrangian:
\begin{equation}
\begin{split}
& {\tilde{L}}_{\text{pair}}  =  \sum_{(i, j) \in \mathcal{S}}  L^S ( D \left(\mathbf{u}_i, \mathbf{u}_j ) \right )
+ \sum_{(i, j) \in \mathcal{D}} L^D ( D  \left(\mathbf{u}_i, \mathbf{u}_j \right) )
 + \psi\sum_{i=1}^N\| \mathbf{u}_i - \mathbf{b}_{i} \|^2_2,\\
& {s.t.} \ \   \mathbf{b}_i, \mathbf{b}_j  \in \{-1, 1\}^{K}, \ \  i,j = 1,...,N,
\end{split}
\label{eqn:pair_Loss_relax2}
\end{equation}
where $\psi$ is the Lagrange multiplier. The term $\sum_{i=1}^N\| \mathbf{u}_i - \mathbf{b}_{i} \|^2_2$ can be viewed as a constraint to minimize the discrepancy between the binary space and the continuous space.

\subsection{Quantized Center Learning}

\begin{figure*}
	\begin{center}
		\begin{tabular}{ccc}
			\includegraphics[width=0.28\textwidth]{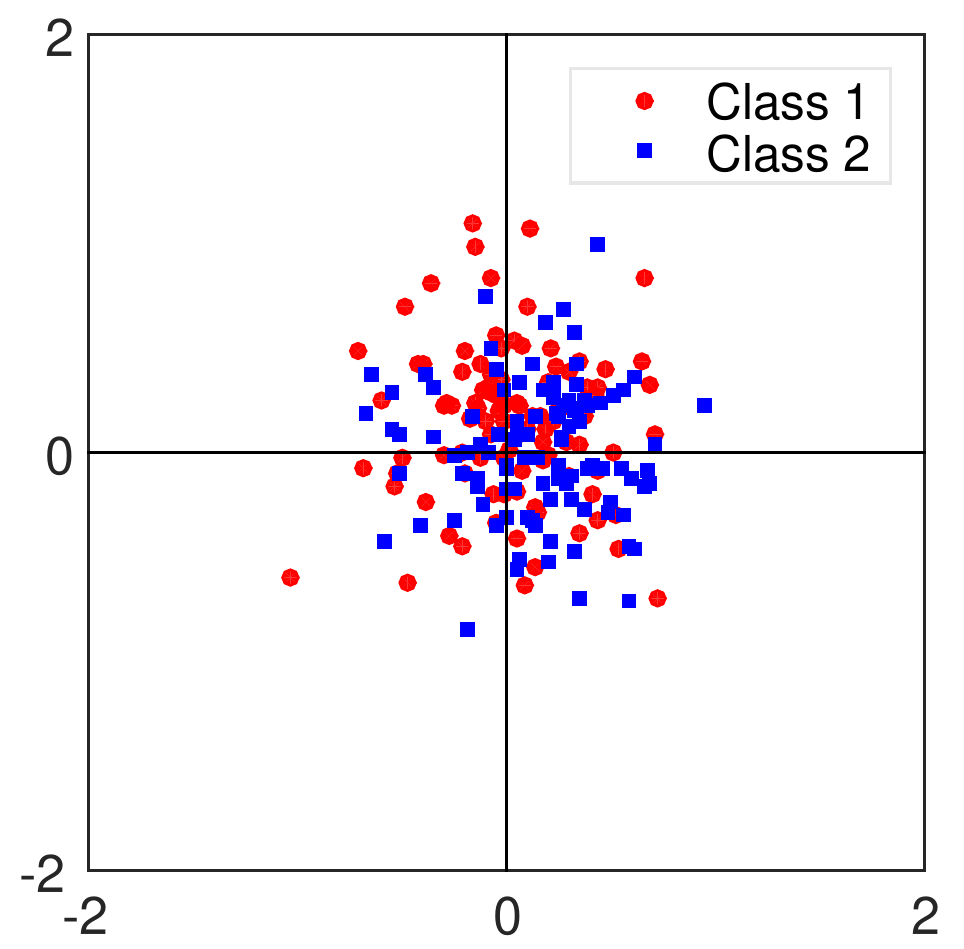} & \ \ \ \
			\includegraphics[width=0.28 \textwidth]{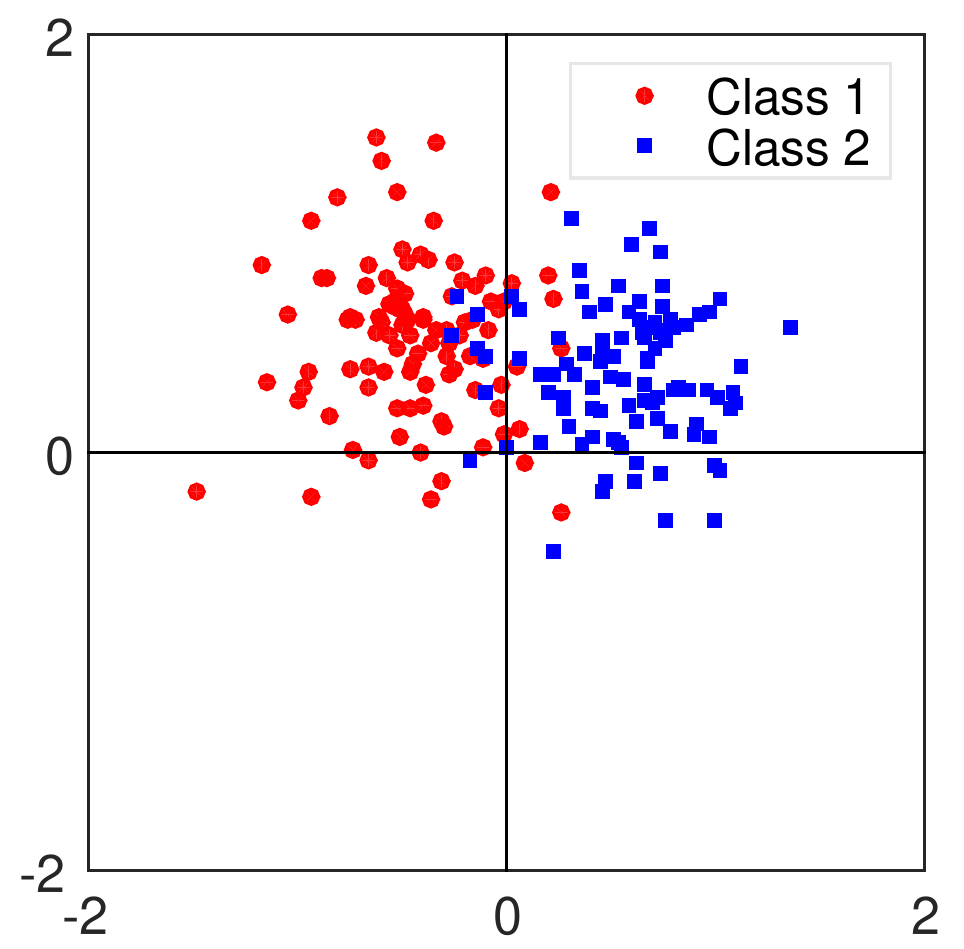} & \ \ \ \
			\includegraphics[width=0.28\textwidth]{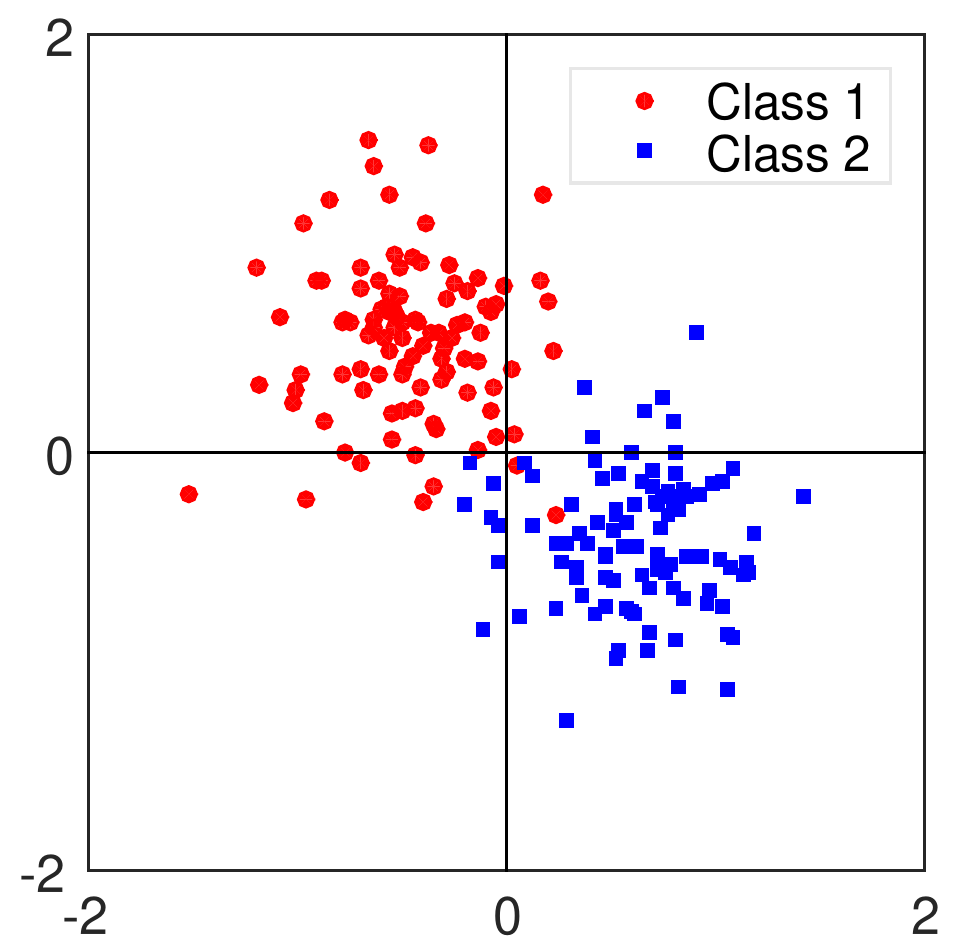} \\
			\ \ \ \ (a): Input & \ \ \ \ \ (b): Only  $L_\text{intra}$  & \  \  \  \ \ (c):  $L_\text{intra}$  + $L_\text{inter}$  \\
		\end{tabular}
	\end{center}
	\caption{  Illustration of Quantized Center learning. All points denote 2D representations extracted by a CNN model from randomly selected two classes samples of CIFAR-10, for 100 samples per class. Binarization is illustrated by quantization $sgn(\cdot)$ (black lines).
		\emph{(a):} Inefficient hashing: Binarization will assign same-class points  to different bins, while assigning different-class points  to the same bins.   \emph{(b):} 
		Using $L_\text{intra}$ clusters classes together and hashing is improved since binarization will assign the classes to different, neighboring bins: class 1 to $[-1,1]$ and class 2 to $[1,1]$.
		\emph{(c):} Using $L_\text{intra}$  + $L_\text{inter}$ also pushes the classes away from each other, improving the hashing further since after binarization class 1 is $[-1,1]$ and class 2 is $[1,-1]$ making the difference between class samples two bit flips.
	}
	\label{fig:shiftVariantLearned}
	\vspace{-0.1in}
\end{figure*}

The Quantized Center Learning module, see \fig{shiftVariantLearned}, maximizes the inter-class distances whilst minimizing the intra-class distances in a quantized setting. To represent class-distances we learn a center for each of the $M$ classes: $\mathbf{C} = \{\mathbf{c}_i\}_{i=1}^M \in \{ -1, 1\}^{K \times M}$, where each center $\mathbf{c}$ is encoded by $K$ bits of binary codes.  Let $\mathbf{u}$ be the network output representation. We then encourage the learned binary code(vertex) of each representation to be close to the corresponding class center while the distance between different class centers is maximized, taking quantization to binary vectors into account.

\textbf{Minimizing intra-class distances ($L_{\text{intra}}$).}  This minimizes the sum of Euclidean distance between the binary codes $\mathbf{b}_i$ of the $N$ training images to their class center:
\begin{equation}
L_{\text{intra}} = \sum_{i=1}^{N} \| \mathbf{b}_i - \mathbf{C} \mathbf{y}_i\|^2_2,
\label{eqn:center_loss1}\vspace{-1mm}
\end{equation}
where all class centers $\mathbf{C}$ are indexed by $\mathbf{b}_i$'s class membership vector $\mathbf{y}_i$.

\textbf{Maximizing inter-class distances ($L_{\text{inter}}$).}
We maximize the sum of pairwise Euclidean distance  between different class centers to maximize the inter-class distance of training data:
\begin{equation}
\sum_{i=1}^{N} \sum_{j=1, j \neq i}^{N} \| \mathbf{c}_i - \mathbf{c}_j \|^2_2 = \sum_{i=1}^{N} \sum_{j=1, j \neq i}^{N} (2K - 2\mathbf{c}_i^\intercal \mathbf{c}_j).
\label{eq:center_dist}
\end{equation}
Since $\mathbf{c}_i, \mathbf{c}_j \in \{-1, 1\}^K$ and $\mathbf{c}_i^\intercal \mathbf{c}_{j\neq i} \ge -K$,  maximizing \eq{center_dist} is equivalent to minimizing
\begin{equation}
 \sum_{i=1}^{N} \sum_{j=1, j \neq i}^{N} (\mathbf{c}_i^{\intercal} \mathbf{c}_j- (-K))^2 = \|\mathbf{C}^{\intercal} \mathbf{C} - K(2I-J_K) \|^2_F,
\label{eq:L2}
\end{equation}
where $\| \cdot \|_F$  denotes the Frobenius norm,  $I$ is the identity matrix and $J_K$ is the all-ones matrix. Simplifying the notation where $A$ replaces $K(2I-J_K)$ yields
\begin{equation}
L_{\text{inter}} = \|\mathbf{C}^{\intercal} \mathbf{C} - A \|^2_F.
\label{eqn:L3}
\end{equation}

\textbf{ Minimizing quantization cost ($L_{\text{quant}}$).} The Center Learning module exploits label information to learn binary codes by minimizing $L_{\mathrm{intra}}$ and $L_{\mathrm{inter}}$ simultaneously. We also need to encourage the learned representation to be close to the quantized binary codes.
$L_{\text{quant}}$ minimizes the total quantization cost in moving representations $\mathbf{u}_i$ towards the desired $\mathbf{b}_i$,
\begin{equation}
L_{\text{quant}} = \sum_{i=1}^{N} \| \mathbf{b}_i - \mathbf{u}_i\|^2_2.
\label{eqn:quant}
\vspace{-0.05in}
\end{equation}

\section{Optimization}

Our proposed Pairwise Similarity module and Quantized Center Learning module are optimized jointly in an alternating fashion where their gradients are back-propagated to train the upstream CNN. Combining the loss functions $\tilde{L}_{pair}$ in~\eqn{pair_Loss_relax2}, $L_{\text{intra}}$ in~\eqn{center_loss1}, $L_{\text{inter}}$ in~\eqn{L3} and $L_{\text{quant}}$ in~\eqn{quant}, the optimization of the whole framework is
\begin{equation}
	\begin{split}
	 \min_{\mathbf{b}_i,\mathbf{u}_i, \mathbf{C}} &\Big[ \varphi \big( \sum_{(i, j) \in \mathcal{S}}  L^S ( D \left(\mathbf{u}_i, \mathbf{u}_j ) \right )
	+ \sum_{(i, j) \in \mathcal{D}} L^D ( D  \left(\mathbf{u}_i, \mathbf{u}_j \right)  )  \big)\\
	 + & \ \mu \sum_{i=1}^{N} \| \mathbf{b}_i - \mathbf{C} \mathbf{y}_i\|^2_2 + \nu \|\mathbf{C}^{\intercal} \mathbf{C} - A \|_F^2 +  \sum_{i=1}^{N} \| \mathbf{b}_i - \mathbf{u}_i\|^2_2 \Big], \\
	\text{s.t.} & \ \   \mathbf{C} \in \{-1, 1\}^{K \times M}, \ \  \mathbf{b}_i \in \{-1, 1\}^K, \ \   i=1,2, \dots, N,
	\end{split}
\label{eqn:loss}
\end{equation}
where $\varphi$, $\mu$ and $\nu$ are hyper-parameters that balance the effect of three objective functions.

Optimizing \eqn{loss} involves the interaction of two types of variables: discrete variables \{$\mathbf{B} = \{\mathbf{b}_i\}_{i=1}^N$, $\mathbf{C}$\} and continuous variables $\mathbf{U} = \{\mathbf{u}_i\}_{i=1}^N$. A typical solution to such multi-variable optimization problem is to alternate between two steps. In particular: 1) optimize $\mathbf{U}$ while fixing $\mathbf{B}$ and $\mathbf{C}$ focusing on $L_{\text{pair}}$ in the Pairwise Similarity Learning module, 2) fixing $\mathbf{U}$ and optimize discrete variables $\mathbf{B}$ and $\mathbf{C}$ in the Quantized Center Learning.

\subsection{Optimizing Pairwise Similarity Learning}

Given $\mathbf{B} = \{\mathbf{b}_i\}_{i=1}^N$, it is straightforward to optimize $\mathbf{U} = \{\mathbf{u}_i\}_{i=1}^N$ by minimizing the subproblem resolved from~\eqn{loss} corresponding to $L_{\text{pair}}$ by gradient descent:
\begin{equation}
\begin{split}
\min_{\mathbf{U}} & \ \ \sum_{i=1}^m \| \mathbf{b}_i - \mathbf{u}_i\|^2_2 + \varphi \big( \sum_{(i, j) \in \mathcal{S}}  L^S ( D \left(\mathbf{u}_i, \mathbf{u}_j ) \right ) + \sum_{(i, j) \in \mathcal{D}} L^D ( D  \left(\mathbf{u}_i, \mathbf{u}_j \right)  ) \big)
\end{split}
\label{eqn:subproblem1}
\end{equation}
Since $\mathbf{U}$ is the output of the last layer of the upstream CNN, which is denoted as $\mathbf{u}_i = \mathbf{W}^\intercal \mathcal{F}_{\text{CNNs}}(\mathbf{x}_i; \mathbf{\theta}) + \mathbf{v}$.
Here $\mathbf{W}$ is the transformation matrix of the last fully connected layer and $\mathbf{v}$ is the bias term. $\mathbf{\theta}$ is the parameters of CNNs before the last layer. For simplicity, we denote all parameters of CNNs models as ${\Theta} = \{\mathbf{W}, \mathbf{v}, {\Theta}\}$.
The CNN parameters are optimized by gradient back-propagation: $\frac{\partial L}{\partial {\Theta}} = \frac{\partial L}{\partial \mathbf{\mathbf{U}}} \frac{\partial \mathbf{U}}{\partial {\Theta}},$
where $L$ is the Loss function corresponding to~\eqn{subproblem1}.

\subsection{Optimizing Quantized Center Learning} \label{Optimizing_Quantized_Center}

With fixed CNN parameters $ {\Theta}$, we learn $\mathbf{B}$ and $\mathbf{C}$ by optimizing the Quantized Center Learning module, as:
\vspace{-0.15in}
\begin{equation}
	\begin{split}
	\min_{\mathbf{B}, \mathbf{C}} & \ \mu \sum_{i=1}^{N} \| \mathbf{b}_i - \mathbf{C} \mathbf{y}_i\|^2_2 + \nu \|\mathbf{C}^{\intercal} \mathbf{C} - A \|_F^2 +  \sum_{i=1}^{N} \| \mathbf{b}_i - \mathbf{u}_i\|^2_2, \\
	\text{s.t.} & \ \   \mathbf{C} \in \{-1, 1\}^{K \times M}, \ \  \mathbf{B} = \{\mathbf{b}_i\}_{i=1}^N \in \{ -1, 1\}^{K \times N}.
	\end{split}
\label{eq:infer}
\end{equation}
We solve this problem by calling alternating optimization strategy again: optimize variables $\mathbf{B}$ and $\mathbf{C}$ by updating one variable with the other fixed.

\noindent \textbf{Initialization of $\mathbf{b}_i$ and $\mathbf{C}$.} Given the representations $\mathbf{u}_i$,
we initialize $\mathbf{b}_i$ as $\mathbf{b}_i= \mathrm{sgn}(\mathbf{u}_i)$.
In the first iteration we initialize the class centers $\mathbf{C}$ with the class mean of the output representations, later we update $\mathbf{C}$ directly.

 \noindent \textbf{Fix $\mathbf{b}_i$, update $\mathbf{C}$.}
 Keeping $\mathbf{b}_i$ fixed in \eq{infer} reduces this sub-problem to
 \begin{equation}
\begin{split}
\min_{\mathbf{C}} & \ \mu \sum_{i=1}^{N} \| \mathbf{b}_i - \mathbf{C} \mathbf{y}_i\|^2_2 + \nu \|\mathbf{C}^{\intercal} \mathbf{C} - A \|_F^2,  \\
 \text{s.t.} & \ \   \mathbf{C} \in \{-1, 1\}^{K \times M}.
\end{split}
\label{eqn:infer_sub1}
\end{equation}
Due to the discrete constraints on the class centers $\mathbf{C}$, the minimization of above problem is a discrete optimization problem which is hard to optimize directly.
We introduce an auxiliary variable $\mathbf{V}$ with the constrain $\mathbf{C} = \mathbf{V}$,
and adding the Lagrange multiplier, the optimization of~\eqn{infer_sub1} is:
\vspace{-0.15in}
\begin{equation}
\begin{split}
& \min_{\mathbf{C}, \mathbf{V}} \ \mu \sum_{i=1}^{N} \| \mathbf{b}_i - \mathbf{V} \mathbf{y}_i\|^2_2 + \nu \|\mathbf{V}^{\intercal} \mathbf{V} - A \|^2_F + \eta \| \mathbf{C}-\mathbf{V} \|^2_F, \\
& \ \text{s.t.} \ \ \quad  \mathbf{C} \in \{-1, 1\}^{K \times M}.
\end{split}
\label{eqn:infer_v}
	\end{equation}

Fixing $\mathbf{V}$, since the optimal solution for $\mathbf{C}$ for
 minimizing $\| \mathbf{C}-\mathbf{V} \|^2_F$ is $\mathbf{C} = \text{sgn} (\mathbf{V})$,
  hence $\| \mathbf{C}-\mathbf{V} \|^2_F$ in~\eqn{infer_v} can be replaced with  $\| \text{sgn}(\mathbf{V})-\mathbf{V} \|^2_F$.
 Let $\mathcal{L}_2$ denote the loss function after applying Lagrange multipliers, then the gradient w.r.t. $\mathbf{V}$ is calculated as:
\begin{equation}
\frac{\partial \mathcal{L}_2}{\partial \mathbf{V}} = 2 \mu (\mathbf{VY}-\mathbf{B})\mathbf{Y}^\intercal + 4 \nu \mathbf{V} (\mathbf{V}^\intercal \mathbf{V}-A) + 2\eta (\mathbf{V} - \text{sgn}(\textbf{V})),
\label{eqn:gd}
\end{equation}
approximating the class center $\mathbf{C}$ with the learned $\mathbf{V}$.

\noindent\textbf{Fix $\mathbf{C}$, update $\mathbf{b}_i$.}
With the variable  $\mathbf{C}$ fixed in~\eq{infer}, we optimize the binary code $\mathbf{b}_i$ with the sub-problem
\begin{equation}
\begin{split}
& \min_{\mathbf{b}_i} \ \mu \sum_{i=1}^{N} \| \mathbf{b}_i - \mathbf{C} \mathbf{y}_i\|^2_2 +  \sum_{i=1}^{N}\| \mathbf{b}_i - \mathbf{u}_i\|^2_2, \\
& \ \text{s.t.}  \ \   \mathbf{b}_i \in \{-1, 1\}^{K}, i=1, \dots, N.
\end{split}
\label{eqn:opti_b}
\end{equation}
We have the closed-form solution of problem (\ref{eqn:opti_b}):
\vspace{-2mm}
\begin{equation}
\mathbf{B} =  \text{sgn} (\mu \mathbf{CY}+ \mathbf{U}).
\label{eqn:opti_finalB}
\end{equation}
See the supplementary for the detailed proof. By defining $\mathcal{F} = \mu \mathbf{CY}+ \mathbf{U}$ as the Fisher's transformed representations, we note that $\mathcal{F}$ is a \emph{translation} transformation of original representations $\mathbf{U}$ which pushes different-class points to different vertex and pulls same-class points to same vertex, while  $\mathcal{F}$ does not change the relative position between same class.
 The learned center $\mathbf{C}$ determines where the corresponding class translates to. The 2D example in
 \fig{shiftVariantLearned} shows that the shape within a class does not change, yet the classes do translate.

\subsection{Joint Optimization}

 We update the two modules jointly, see supplementary material.
In each iteration, the Pairwise Similarity Learning module and Quantized Center Learning module are optimized in an alternating way to learn the continuous variable $\mathbf{U}$ and discrete variables \{$\mathbf{B}, \mathbf{C}$\}, respectively.

\section{Experiments}

\textbf{Datasets.} We conduct experiments on three datasets: CIFAR-10, NUS-WIDE and ImageNet100.
{CIFAR-10} consists of $60$k  color images with the resolution of $32 \times 32$ categorized into 10 classes. Each image has a single label. {NUS-WIDE} is a multi-label dataset, which contains 269,648 color images collected from Flickr. There are 81 classes, where each image is annotated with one or multiple class labels. Following~\cite{LaiCVPR2015,Li2017NIPS, Liu11hashingwith}, we use  a subset of 195,834 images associated with 21 most frequent classes (concepts) for evaluation, among which 105,972 images has more than two labels and 89,862 images have a single label. Each class contains at least 5,000 samples. {ImageNet100} consists of 130K single labelled images from 100 categories, which is a subset of the large benchmark ImageNet~\cite{Deng2009ImageNet}.

\noindent
\textbf{Experimental settings.}
 Following~\cite{Li2017NIPS, LiIJCAI2016}, 100 random images per class in CIFAR-10  form the test query set and 500 images per class are the training set. For NUS-WIDE, we randomly select 100 images per class as test queries and 500 images per class  as the training set. The pairwise ground truth for two images sharing at least one common label is similar and otherwise dissimilar. Following~\cite{cao2017hashnet}, we sample 100 images per class for ImageNet100 to construct a training set,  and all the images in the validation set are used as the test set.

\noindent \textbf{Evaluation metrics.}
We evaluate retrieval performance using: mean Average Precision (MAP),
precision of the top N returned examples (P@N), Precision-Recall curves (PR) and Recall curves (R@N).
 All compared methods use identical training and test sets for fair comparison.
 For NUS-WIDE, we adopt MAP@5000 and MAP@50000 for the small-data  setting and large-data  setting, respectively. We show the results of MAP@1000 for ImageNet100.

\noindent \textbf{Network and parameter settings.}
To have a fair comparison with previous methods~\cite{Li2017NIPS, LiIJCAI2016, wang2016deep}, we fine-tune the VGG-F\cite{LiIJCAI2016, Li2017NIPS} architecture for the experiments on CIFAR-10 and NUS-WIDE while the AlexNet architecture~\cite{Alexnet} is fine-tuned for the experments on ImageNet100.
Both deep network architectures are pre-trained on ImageNet.
The hyper-parameters $\{\varphi, \mu, \eta, \nu,\}$ are tuned by cross-validation on a validation set  and the margin $m$ is chosen from $\{0.5, 1, 1.5, 2\}$. Stochastic Gradient Descent (SGD) is used for optimization.

\subsection{Exp 1: Effect of Quantized Center Learning}

To investigate the effect of $L_{\text{Intra}}$ (minimizing intra-class distances) and $L_{\text{Inter}}$ (maximizing inter-class distances) in the Quantized Center Learning module, we conduct an ablation study in the small-data setting which starts with the Pairwise Similarity Learning module $L_{\text{pair}}$ in~\eqn{pair_Loss_relax2} in the model and then augment the model incrementally with $L_{\text{intra}}$ in~\eqn{center_loss1} and $L_{\text{Inter}}$ in~\eqn{L3}. In~\tab{centerlearning} we show the experimental results. We observe that both $L_{\text{Intra}}$ and $L_{\text{Inter}}$ contribute substantially to the performance of the whole model.

\begin{table}
	\centering
	\renewcommand{\arraystretch}{1.0}
	\resizebox{0.7\linewidth}{!}{
	\begin{tabular}[c]{ccccccc}
		\toprule
		 & \multicolumn{2}{c}{Components} & \multicolumn{2}{c}{CIFAR-10} & \multicolumn{2}{c}{ImageNet100} \\ 
		Baseline & \textbf{$L_{\text{Intra}}$} & \textbf{$L_{\text{Inter}}$} & 12 Bits &  24 Bits & 16 Bits & 48 Bits  \\ \midrule
		
		& $\times$ &$\times$ &0.730 & 0.787  & 0.431 &0.572  \\
		$L_{\text{pair}}$  &\checkmark  & $\times$  &0.746 &  0.802 & 0.543   &   0.696 \\
		& \checkmark &\checkmark &0.772 & 0.809 & 0.576 &  0.726 \\
		\bottomrule
	\end{tabular}
	}
\vspace{0.05in}
	\caption{Comparative results for our model with different components of the Quantized Center Learning module on CIFAR-10 and ImageNet100 .
	We start with the Pairwise Similarity Learning ($L_{\text{pair}}$) and augment incrementally with two components: $L_{\text{Intra}}$ in~\eqn{center_loss1} and $L_{\text{Inter}}$ in~\eqn{L3}. For 24-bits in CIFAR-10 the performance seems already saturated; for all other settings, each added component brings an advantage. }
	\label{tab:centerlearning}
		\vspace{-0.1in}
\end{table}

\subsection{Exp 2: Functionality of different modules}

\begin{figure}[!t]
	\begin{center}
		$\begin{array}{cccc}
		\includegraphics[width=0.23\linewidth]{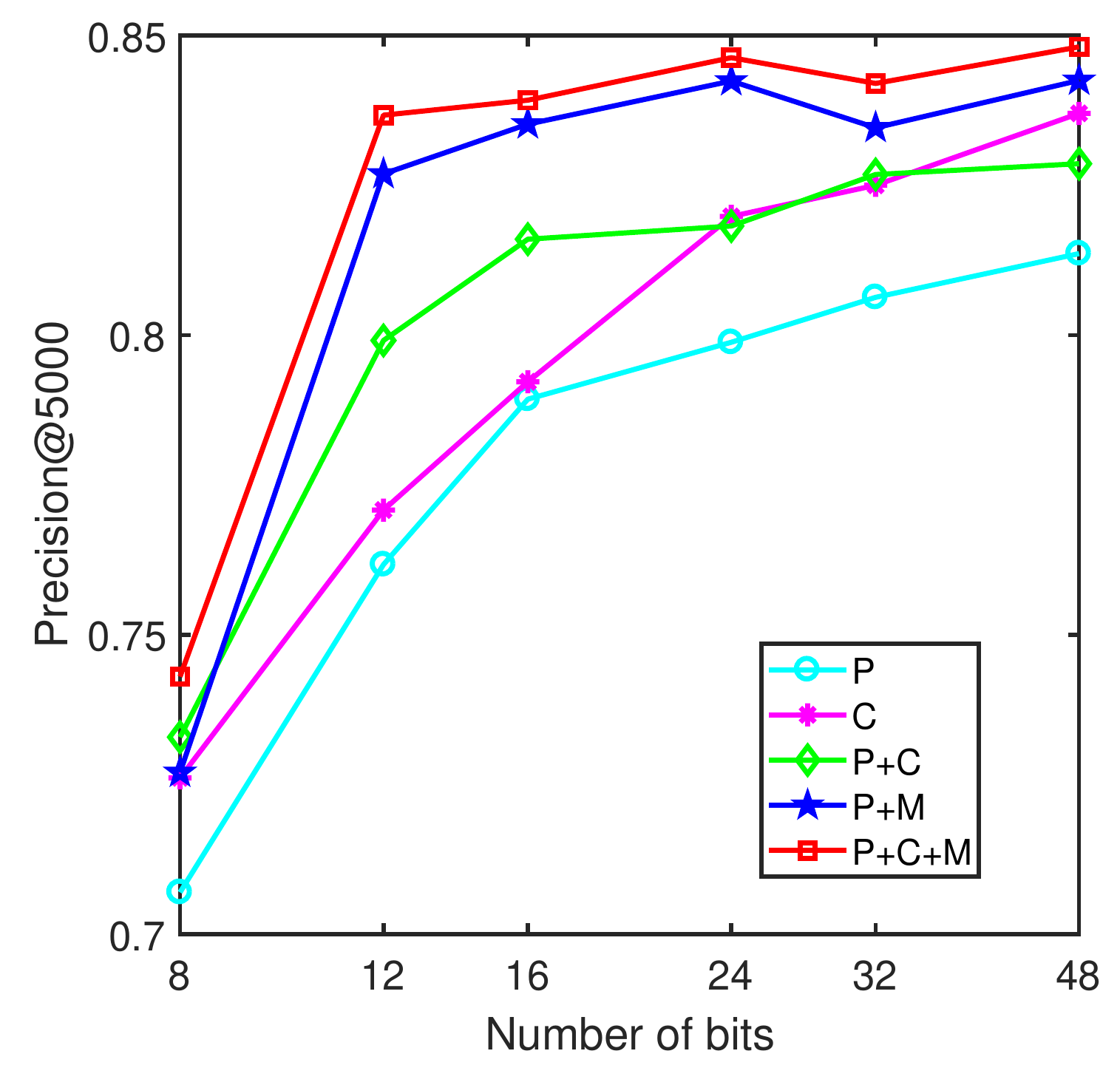} &
		\includegraphics[width=0.23\linewidth]{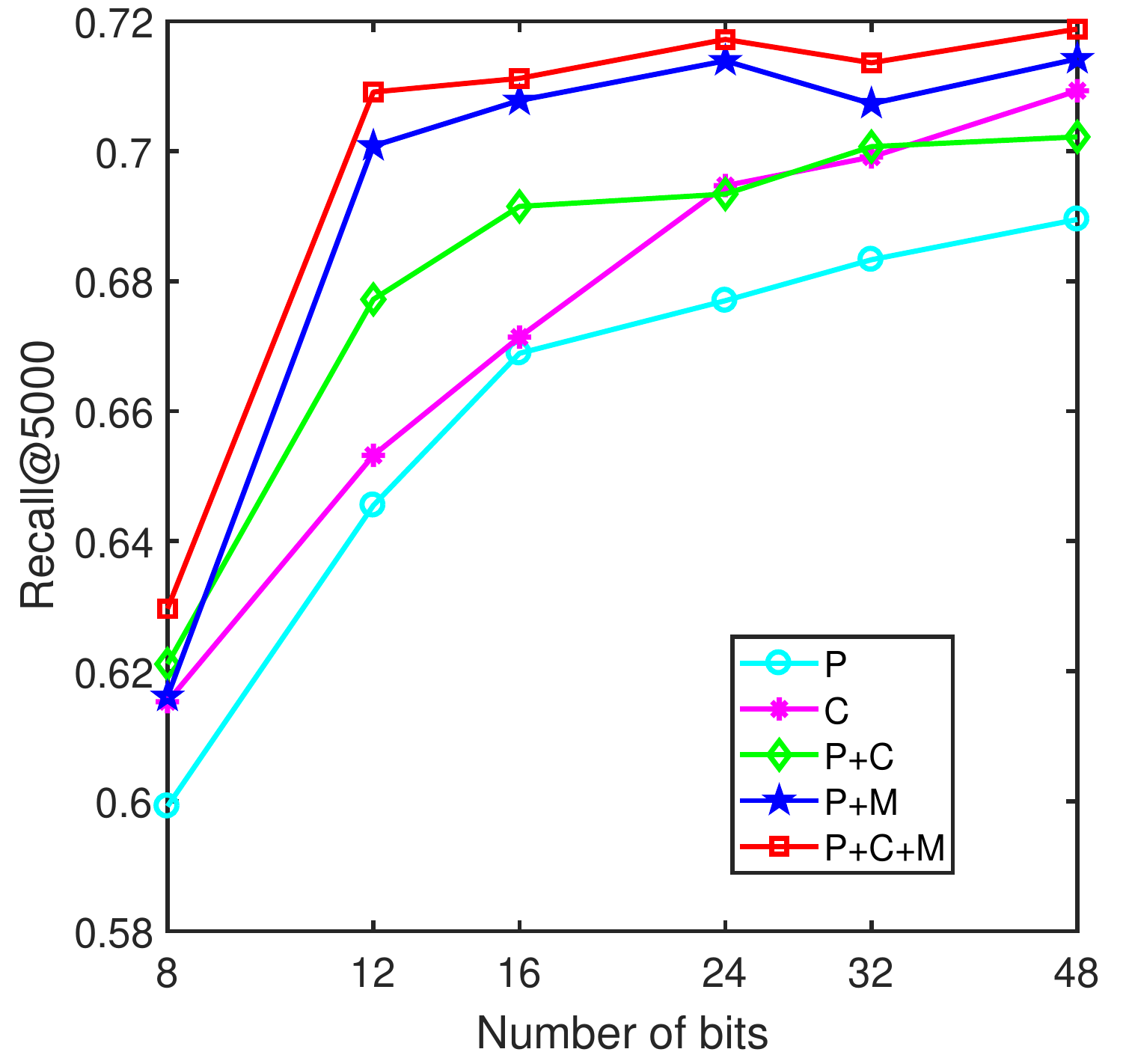}  &
		\includegraphics[width=0.23\linewidth]{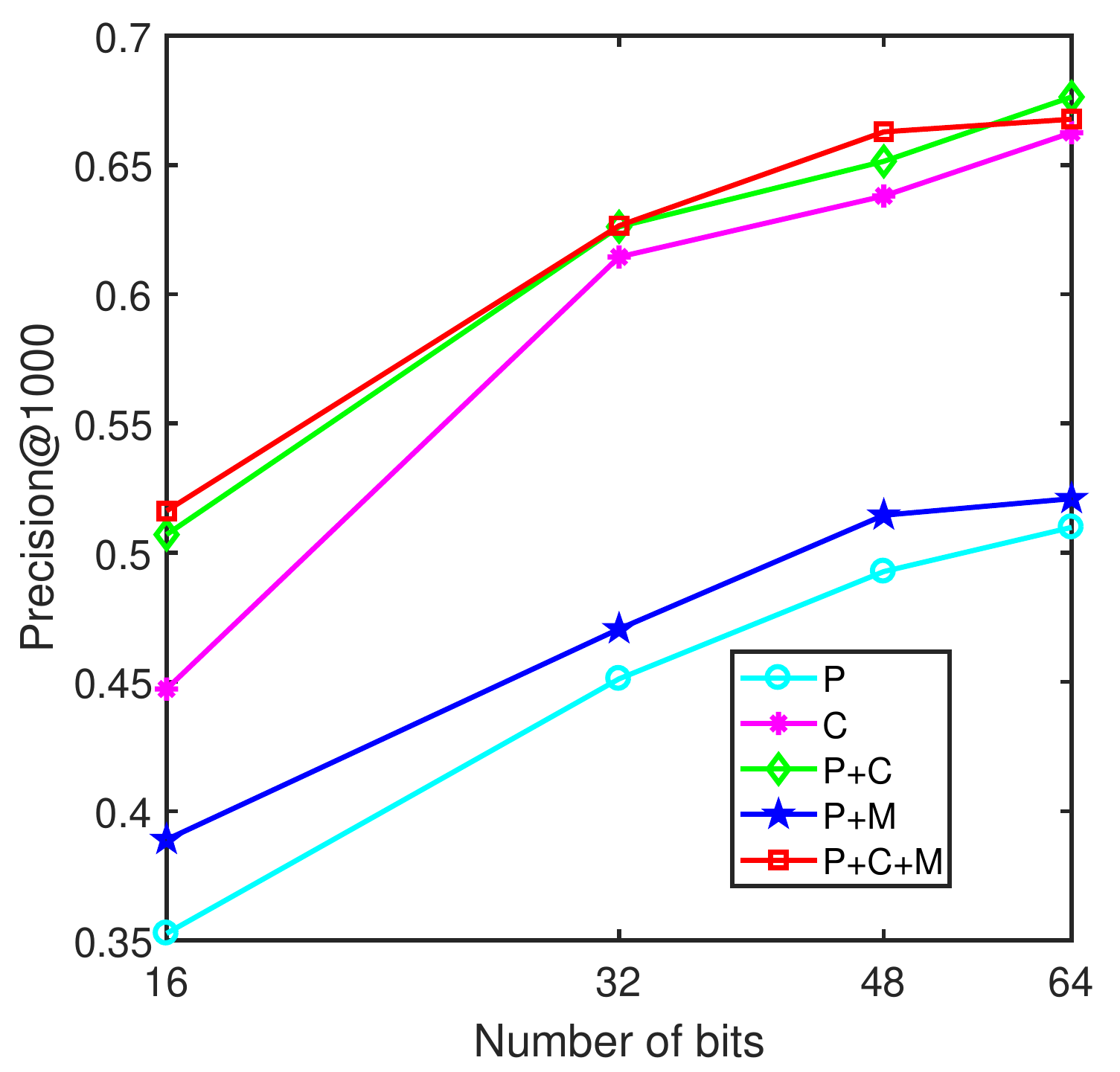} &
		\includegraphics[width=0.23\linewidth]{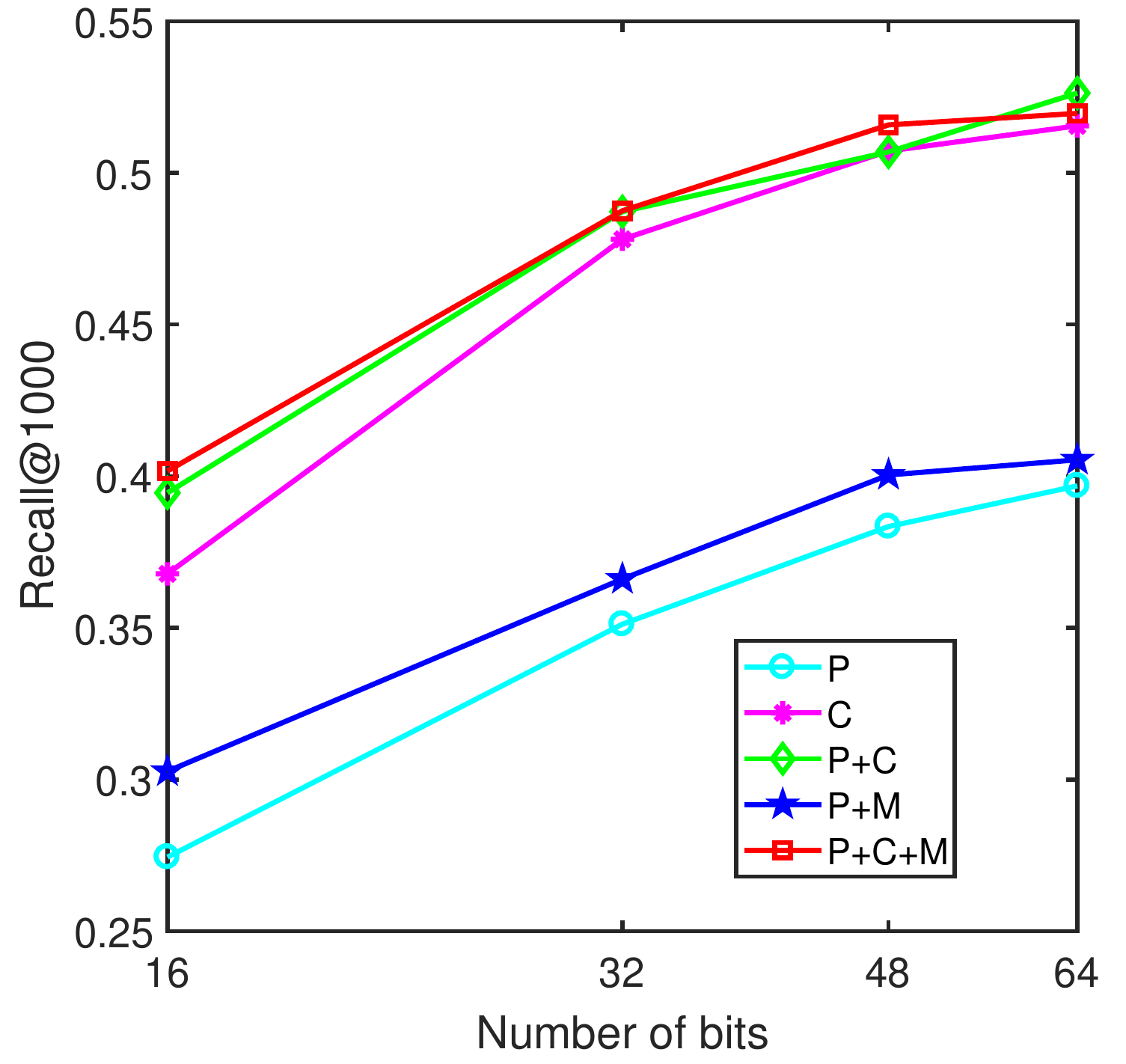} \\		
			\multicolumn{2}{c}{\text{CIFAR-10}}  &	 \multicolumn{2}{c}{\text{ImageNet100}} \\
		\end{array}$
	\end{center}
	\caption{Evaluating different modules on two datasets. Herein $\textbf{P}$ refers to the Pairwise Similarity Learning module without margin while $\textbf{C}$ refers to the Quantized Center Learning module. $\textbf{P}+\textbf{M}$ denotes the Pairwise Similarity Learning module with tuned margin.}
	\label{fig:diff_modules}
\end{figure}

We evaluate the effect of combining modules on both  CIFAR-10 and ImageNet100 datasets using precision and recall curves for top 5,000 returned images for different number of bits. In~\fig{diff_modules} we compare on CIFAR-10 and ImageNet100. We observe that each module adds value. The only exception is Fisher-only, which outperforms the combined Pairwise+Fisher model for a code size of 48. Second, the combined models can get relatively well for fewer bits, while the single models need more bits to achieve the same performance.

The results on ImageNet100 shown in \fig{diff_modules} indicate that the Quantized Center Learning module improves the performance substantially. One potential explanation is that the Pairwise Similarity Learning module ($\tilde{L}_{\text{pair}}$) is sensitive to the balance between the positive and negative training sample pairs, which is hard to achieve in the data with large number of classes.
In contrast, the Quantized Center Learning module does not suffer from this limitation. The sensitivity of the margin $m$ is in  the supplemental.

\subsection{Exp 3: Comparison with others }

\begin{table*}
	\centering
	\footnotesize
	\renewcommand{\arraystretch}{1.0}
	\resizebox{1\linewidth}{!}{
		\begin{tabular}[c]{lcccclcccc}
			\toprule
			\multirow{2}*{Method} & \multicolumn{4}{c}{CIFAR-10} & \multirow{2}*{Method} & \multicolumn{4}{c}{NUS-WIDE}\\
			& 12 bits & 24 bits  & 32 bits  & 48 bits &  &  12 bits & 24 bits  & 32 bits  & 48 bits \\
			\midrule
			Ours &\textbf{0.803}&\textbf{ 0.825} & \textbf{0.831}& \textbf{0.844} & Ours &\textbf{0.795}&\textbf{ 0.823} & \textbf{0.833}& \textbf{0.842} \\
			DSDH~\cite{Li2017NIPS}& 0.740 & 0.786 & 0.801 & 0.820 & DSDH~\cite{Li2017NIPS}& 0.776 &0.808 &0.820 &0.829 \\
            Greedy Hash~\cite{su2018greedy}& 0.774   & 0.795 & 0.810 & 0.822 &  Greedy Hash~\cite{su2018greedy} & -- &-- &-- &-- \\
			DPSH~\cite{LiIJCAI2016} &0.713 &  0.727& 0.744 & 0.757 &  DPSH~\cite{LiIJCAI2016} &0.752 &0.790 &0.794 &0.812  \\
			DQN~\cite{CaoL0ZW16} &  0.554 & 0.558 & 0.564 & 0.580  &  DQN~\cite{CaoL0ZW16} & 0.768 &0.776 &0.783& 0.792   \\
			DTSH~\cite{wang2016deep}  &0.710 &0.750 &0.765 &0.774 & DTSH~\cite{wang2016deep}  &0.773& 0.808& 0.812 &0.824 \\
			NINH~\cite{LaiPLY15} &0.552 &0.566 &0.558 &0.581 &  NINH~\cite{LaiPLY15} &0.674 &0.697& 0.713 &0.715 \\
			CNNH~\cite{Xia:2014:AAAI} &0.439 &0.511 &0.509 &0.522 & CNNH~\cite{Xia:2014:AAAI} &0.611 &0.618 &0.625 &0.608 \\
			\bottomrule
		\end{tabular}

	}
	\vspace{0.001in}
	\caption{MAP for various methods for the small-data setting for CIFAR-10 and NUS-WIDE. The best performance is boldfaced. For NUS-WIDE, the top 5,000 is used for the MAP.}
			\label{table:experiment_first}
\end{table*}

In~\tabl{experiment_first} we show results on both CIFAR-10 and NUS-WIDE datasets in the small-data setting. In particular for a few number of bits, our model compares well to others. It is worth noting that the performance  comparison among
VGG-F and AlexNet networks is considered to be fair~\cite{su2018greedy}, since both architectures have the same network composition.

The state-of-the-art DSDH~\cite{Li2017NIPS} model also uses pairwise labels and classification labels. The major difference between is in using the classification label: DSDH~\cite{Li2017NIPS} learns hash codes by maximizing the classification performance while our model learns centers to model between-class and between-sample distances. While DSDH performs excellent, our model outperforms DSDH in all experiments.

Another interesting observation is that SDH~\cite{Shen2015Supervised}, which is based on sole classification label information, performs competitively on NUS-WIDE but not as good on CIFAR-10. In contrast, our model and DSDH~\cite{Li2017NIPS} that leverage two types of information, perform much more robust. It reveals the necessity of incorporating the pairwise label information.

We also conduct experiments to compare our method to other baseline models on ImageNet100 and the results are presented in \tabl{imageNet}. It is observed that our model achieves the best performance on all bits except for the $16$ bits.

\begin{table}[htb]
				\vspace{-0.05in}
	\centering
	\renewcommand{\arraystretch}{1}
	\resizebox{0.56\linewidth}{!}{
	\begin{tabular}[c]{lcccc}
		\toprule
		 & \multicolumn{4}{c}{ImageNet100 (mAP@1K)}\\ 
		Method& 16 Bits & 32 Bits & 48 Bits  & 64 Bits   \\
		\midrule
        CNNH~\cite{Xia:2014:AAAI}  & 0.281 & 0.450 & 0.525 & 0.554 \\
        NINH~\cite{LaiPLY15}  & 0.290 & 0.461 & 0.530 & 0.565 \\
        DHN~\cite{Zhu:2016:AAAI}& 0.311 & 0.472 & 0.542 & 0.573 \\
		HashNet~\cite{cao2017hashnet}& 0.506 & 0.630 & 0.663 & 0.683  \\
		Greedy Hash~\cite{su2018greedy} & \textbf{0.625} & 0.662 & 0.682 & 0.688 \\
        Ours  & {0.590} & \textbf{0.697} & \textbf{0.726} & \textbf{0.747} \\
		\bottomrule
	\end{tabular}
	}

	\vspace{0.05in}
		\caption{MAP@1K results on ImageNet100 using AlexNet.}
								\label{table:imageNet}
					\vspace{-0.2in}

\end{table}

\section{Conclusion}

We present a supervised deep binary hashing method focusing on binary separability through a pair-wise margin and inspired by Fisher's linear discriminant which minimizes within-class distances while maximizing between-class distances. For medium-sized datasets with much training data --where larger hash codes can be used-- our method performs on par or only slightly better than other methods. Our method is most suitable for extremely large datasets with few training data where only tiny bit codes can be used; there our method compares most favorably to others.

\clearpage

{\footnotesize
\bibliographystyle{ieee}
\bibliography{main}
}
\end{document}

%% file: macrosjvg.tex
\usepackage{xcolor}

\newcommand{\Eq}[1]{Eq.~(\ref{eq:#1})}
\newcommand{\eq}[1]{\Eq{#1}}
\newcommand{\Eqn}[1]{Eq.~(\ref{eqn:#1})}
\newcommand{\eqn}[1]{\Eqn{#1}}
\newcommand{\fig}[1]{Fig.~\ref{fig:#1}}
\newcommand{\tab}[1]{Table~\ref{tab:#1}}
\newcommand{\tabl}[1]{Table~\ref{table:#1}}

\usepackage{graphicx}
\usepackage{amsmath}
\usepackage{amssymb}
\usepackage{booktabs}

\usepackage{xspace}
